**Accepted for publication in *Nature Machine Intelligence***

# Tensegrity Continuum Robots Enable Task-Adaptive Morphologies for Cooperative Behaviors

**Authors:**

M. H. Saikot[1]†, S. Spiegel[1]†, S. A. Kariyawasam[1], A. Stefka[1], J. Chrisler[1], J. Zhao[1]*

**Affiliations:**

[1]Adaptive Robotics Lab, Department of Mechanical Engineering, Colorado State University, Fort Collins, CO, USA

*Corresponding author. Email: Jianguo.Zhao@colostate.edu.

† These authors contributed equally to this work.

**Abstract:** Robots that can change their morphologies and behaviors for different tasks and environments hold great promise for adaptable, multifunctional systems. Modular reconfigurable robots (MRRs) can achieve such functionalities by docking and rearranging individual units, but most rely on rigid modules that lack structural compliance, resulting in limited capabilities. Continuum robots offer compliance through flexible backbones, yet they cannot self-reconfigure into task-adaptive multi-robot configurations. Here, we introduce an MRR that unifies the advantages of both architectures by combining a tensegrity-based compliant body with claw-based connection mechanisms. Each robot can manipulate and locomote independently, and multiple robots can self-reconfigure into different morphologies (e.g., chains, loops, branches) for cooperative manipulation and locomotion. We demonstrate the robots' capability across diverse tasks and environments, including coordinated object manipulation and transport, multimodal locomotion, and loco-manipulation in real-world scenarios. These results lay a foundation for adaptable and multifunctional robotic collectives, with broad potential applications in manufacturing, space exploration, and search-and-rescue operations.

## INTRODUCTION

Robots that can reconfigure their morphology for different tasks and environments offer greater functional adaptability[1–3]. Among these, modular reconfigurable robots (MRRs) can use a fixed set of modules to dock and rearrange into different morphologies. MRR modules have been realized in various structural forms, including cuboidal blocks[4–9], spherical modules[10–12], cylindrical bodies[13,14], and triangular units[15], among others. They can be interconnected to form different morphologies, such as chains[6,16–18], lattices[19,20], truss-based[21], mobile platforms[8,22], or free-form assemblies[10–12,23]. The diverse morphologies enable MRRs to perform a wide range of tasks, including manipulation[24,25], diverse modes of locomotion (e.g., crawling[20], rolling[26], walking[27], snake-like[16], slip-sticking[28], legged-locomotion[29,30], swimming[31]), object transportation[32], and combinations of these capabilities[19,33–37]. Owing to their adaptive capabilities, MRRs have demonstrated significant potential across diverse application domains[38], including space operations[39], underwater exploration[24,31,40], aerial assembly[22], and construction[7,41]. However, existing MRRs are largely built from rigid cuboidal[4,8,42] or spherical modules[10,11] that achieve shape change through docking, folding, or rearrangement instead of body compliance. Without compliance, these modules generally remain rigid bodies after connection with limited ability to conform to objects, terrain, or docking misalignment, restricting contact-rich interaction with the environment and with other modules. As a result, these systems generally excel at either manipulation or locomotion, but rarely both, leaving a critical gap in the combined loco-manipulation capabilities.

In contrast, continuum robotic systems provide the body compliance that rigid MRRs lack. With flexible or extensible backbones, their inherent compliance makes them well suited for constrained and complex environments[43–45]. These systems have been realized through a variety of mechanisms, such as flexible spines[46–48], origami structures[49–51], and soft bodies[24,35,52,53]. They also have different actuation methods, including tendons[46–48,51,54,55], external magnetic field[49,56,57], pneumatics[24,35,50,52,58], shape memory alloy[59,60], etc. However, most existing continuum robots function as standalone systems rather than self-contained modules that can dock, detach, and reassemble into multi-robot configurations. As a result, it remains largely unexplored how continuum-based designs can be leveraged for new MRRs that combine compliance with modular reconfiguration to achieve task-adaptive morphologies for locomotion and manipulation across diverse environments.

In this paper, we present TeCoBot (Tensegrity-based Continuum robot), a modular self-reconfigurable robot that addresses key limitations of existing MRRs by combining body compliance with modular reconfigurability. Unlike existing tensegrity robots, which typically actuate a single tensegrity unit primarily for locomotion[61–65], TeCoBot stacks multiple tensegrity modules into a compliant continuum body, enabling both axial collapse/extension and omnidirectional bending (Fig. 1a(i)). Tensegrity combines large recoverable deformation with load-bearing stiffness: rigid struts carry compression and transmit load, while elastic cables provide tension, compliance, and recovery. This combination reflects a form of embodied intelligence[66], in which compliance offloads part of the sensing and control burden onto the body itself. Each TeCoBot is equipped with claws at both ends, which serve as grasping tools and connection interfaces to other TeCoBots. This distinctive design allows an individual TeCoBot to manipulate (Fig. 1a(ii)), locomote in diverse modes (e.g., axial extension–compression, rolling, and pipe climbing, Fig. 1a(iii)), as well as integrate these capabilities for loco-manipulation (Fig. 1a(iv)). TeCoBot exhibits strong performance compared to other representative modular robots in both manipulation and locomotion (Fig. 1b), with detailed comparisons and rationale provided in Supplementary Note 1 and Tables S1–S3.

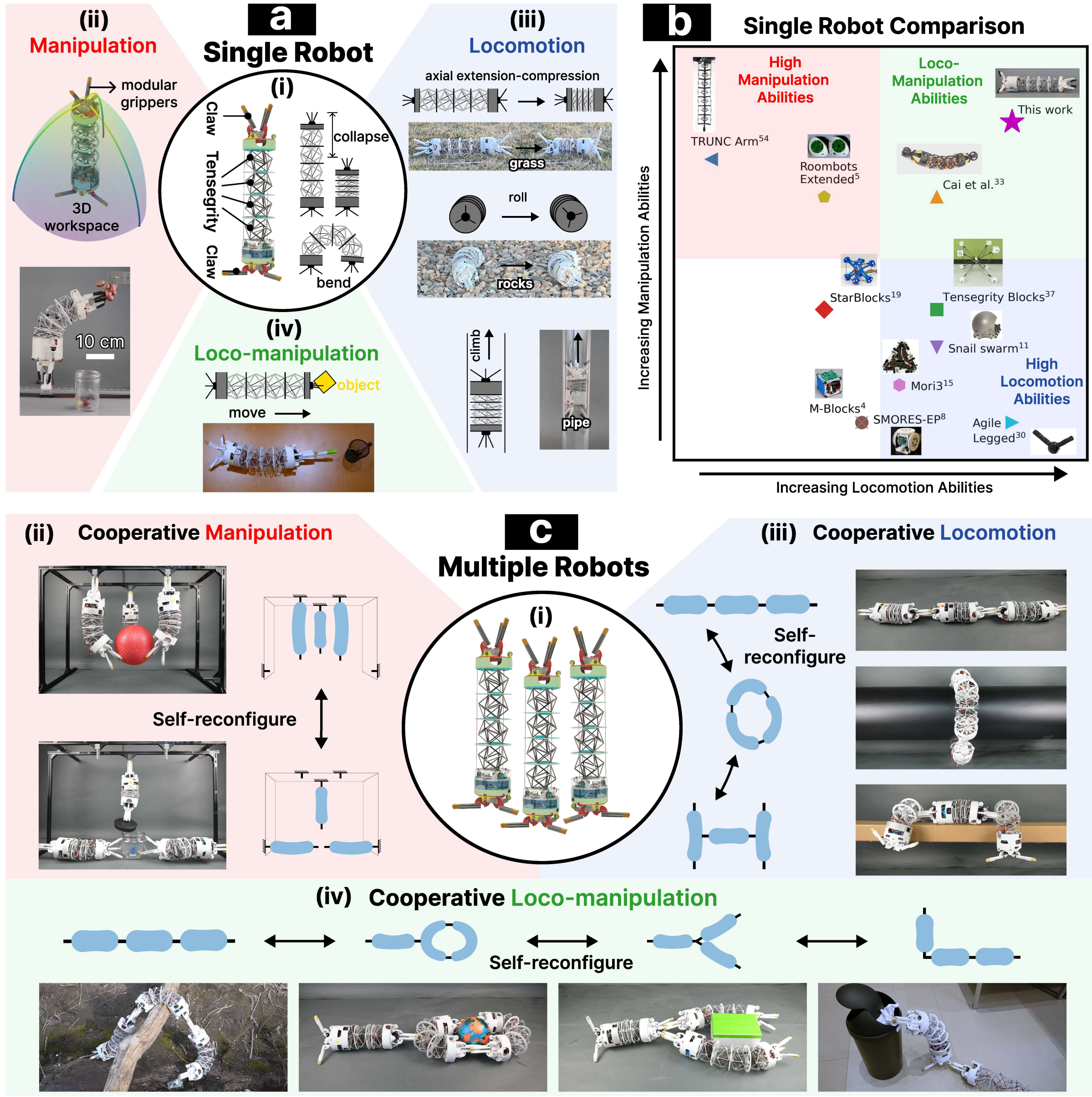

**Figure 1: Overview of TeCoBot's individual and multi-robot capabilities enabled by its tensegrity-based continuum body and claw-based connection mechanisms. (a)** Single-robot capabilities. *(i)* A single TeCoBot performs axial collapsing/extension and omnidirectional bending through cable-driven actuation of its stacked tensegrity modules. *(ii)* Manipulation enabled by body deformation and dual modular claws allows operation within a 3D workspace. *(iii)* Locomotion capabilities include using axial extension–contraction, rolling, and pipe climbing, among others. *(iv)* Loco-manipulation integrates manipulation and movement, allowing the robot to transport objects while repositioning itself. **(b)** Comparison of single-robot capabilities with representative MRRs. TeCoBot occupies a previously unfilled region of the manipulation–locomotion capability space. **(c)** Multi-robot capabilities enabled by self-reconfiguration. *(i)* Three TeCoBots can self-reconfigure using claw-to-claw and claw-to-body connections to form different morphologies. *(ii)* Cooperative manipulation: multiple robots reconfigure their morphology to grasp and manipulate large or complex objects. *(iii)* Cooperative locomotion: robots coordinate anchoring and actuation to locomote as reconfigurable assemblies across different environments. *(iv)* Cooperative loco-manipulation: integrated multi-robot behaviors enable complex sequences such as climbing, object transport, and handover through dynamic reconfiguration.

Beyond single-robot behaviors, multiple TeCoBots can self-reconfigure into task-adaptive morphologies for cooperative behaviors. For instance, when docked to a frame, TeCoBots can self-reconfigure for cooperative manipulation tasks, such as lifting a large ball or removing a container lid (Fig. 1c(ii)). In different environments, they can self-reconfigure into different morphologies for cooperative locomotion, such as axial extension–compression, loop rolling around pipes, and duct climbing (Fig. 1c(iii)). These capabilities can be further combined for cooperative loco-manipulation, including traversing a log while carrying litter, transporting a payload within a loop, moving a large box in a Y-assembly, and disposing trash into a bin (Fig. 1c(iv)). Overall, TeCoBot demonstrates how continuum-based body compliance can be incorporated into MRRs to form task-adaptive morphologies for cooperative behaviors.

## RESULTS

### Design of TeCoBot

Each TeCoBot consists of a compliant and continuum body made of multiple tensegrity modules (Fig. 2a). One end of the body connects to a base module that houses all the electronics, while the other end connects to a top module, with a claw mounted at each end. The robot measures 587 mm (length) by 110 mm (diameter) and weighs ~600 g (excluding batteries).

The robot's body is composed of several vertically stacked icosahedral tensegrity modules (Fig. 2b). Each module is made from six rigid carbon fiber rods and a continuous stretchable tendon, which is 3D printed as a flat lattice structure using thermoplastic polyurethane (TPU) (top of Fig. 2b). The tendon is assembled onto the rods to form an icosahedron tensegrity[61,62] that can bend omnidirectionally and collapse vertically. Adjacent tensegrity modules are connected using a connector piece (Fig. 2b) mounted onto 3D-printed spacer disks (Fig. 2a), which ensure structural stability and guide the actuation cables. Three cables spaced 120° apart are routed through the spacer disks along the body's longitudinal axis, enabling omnidirectional bending through cable displacement control.

The base module (Fig. 2c) houses the cable-actuation motors and encoders for closed-loop cable control (see Methods for component details). The top module houses the main battery that powers all the actuators. Both ends of the body carry modular, swappable claws (Fig. 2d). In our experiments, we use two interchangeable claw designs: a 3D-printed triangular claw with rubber sleeves for improved locomotion and multi-point docking, and a flexible TPU-padded claw for delicate object handling. Each claw is actuated by a single motor through a self-locking worm gear, ensuring secure grasping and stable docking. Passive roller wheels (Fig. 2a) are placed between the claw fingers to facilitate smooth ground contact during locomotion.

TeCoBots communicate wirelessly via an onboard ESP32 receiver module (Fig. 2e). A PC-connected ESP32 transmitter sends commands wirelessly to each robot's onboard ESP32 receiver, which controls the motors through dedicated motor drivers. This setup supports both independent and synchronized control of multiple connected robots.

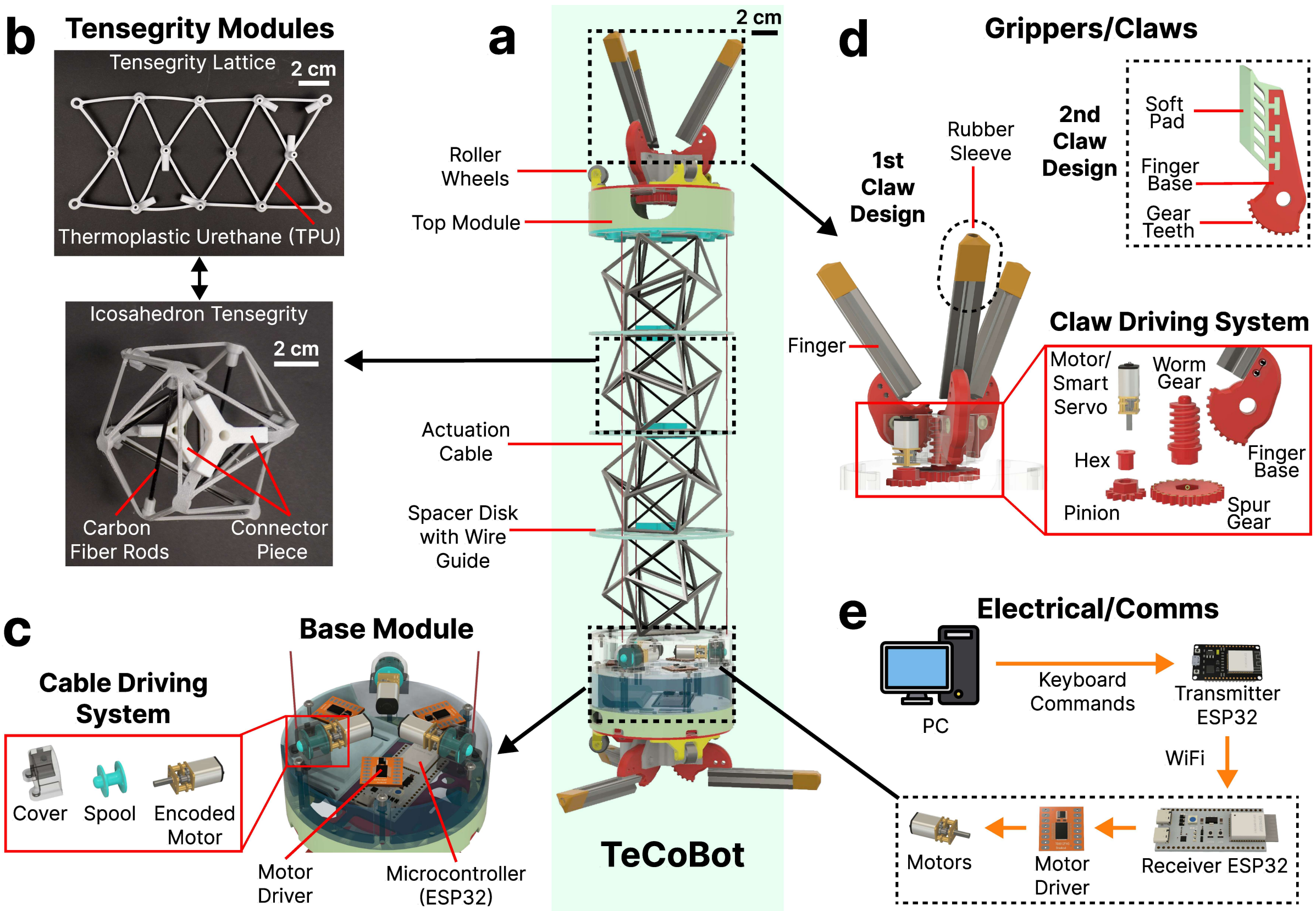

**Figure 2: Design and components of the TeCoBot. (a)** Overall robot architecture. TeCoBot consists of vertically stacked tensegrity modules forming a compliant continuum backbone actuated by three longitudinal cables. Spacer disks guide cable routing, while dual end-claws and integrated roller wheels support grasping and locomotion. **(b)** Tensegrity module structure. Each module is constructed from six carbon-fiber rods and a continuous stretchable thermoplastic polyurethane (TPU) tendon lattice assembled into an icosahedral tensegrity geometry. Connector pieces join adjacent modules and interface with spacer disks to maintain structural alignment and cable pathways. **(c)** Cable-driven actuation system. Three geared DC motors housed in the base module drive spools for bidirectional cable control. Each motor is equipped with an encoder and controlled by an onboard ESP32 microcontroller through dedicated motor drivers. **(d)** Claw design and driving mechanism. Two interchangeable claw designs are used: a rubber-sleeved triangular claw for locomotion and multi-point docking, and a soft-pad claw for manipulating delicate objects. Each claw is actuated by a motor–worm-gear system that provides self-locking, transmitting torque through hex, pinion, and spur gears to drive the three claw fingers. **(e)** Wireless communication and control. A PC-connected ESP32 transmitter sends Wi-Fi commands to each robot's onboard ESP32 receiver, which drives the motors and enables both independent and synchronized control of multiple robots.

## Characterization

To inform the design of TeCoBot, we characterize how TPU tendon's cross-sectional geometry and number of tensegrity modules influence deformation capability and force requirements (Fig. 3a-f). We further evaluate the claw's docking force across sleeve materials, lengths, and gripping forces (Fig. 3g-i). We also compare kinematic model predictions with experimentally reconstructed robot shapes (Fig. 3j-l).

We first examine axial collapsing of a single tensegrity module using tendons with three cross-sectional sizes (1 × 2 mm, 2 × 2 mm, and 3 × 2 mm). Increasing tendon thickness significantly raises the peak collapsing force from 17.5 ± 1.0 N (1 × 2 mm) to 27.4 ± 0.8 N (2 × 2 mm), with a small increase for larger tendon cross sections (30.9 ± 1.0 N for 3 × 2 mm, Fig. 3a). Based on this trade-off between rigidity and actuation effort, we select 2 × 2 mm tendons for TeCoBot. Using this tendon dimension, we further investigate the effect of stacking multiple modules. The maximum axial displacement increases linearly with the number of modules, from 60 mm for one to 260 mm for four modules (Fig. 3b), while the peak collapsing force remains approximately constant (~23–28 N, Fig. 3c). This demonstrates a key advantage of stacking multiple modules: large axial deformations can be achieved through structural scaling without increasing actuation force.

We perform a similar characterization for bending (Fig. 3d–f). For a single tensegrity module, the bending force rises linearly with the bending angle, and thicker tendons require higher peak bending forces, from 11.1 ± 0.9 N for 1 × 2 mm and 19.0 ± 0.7 N for 2 × 2 mm, to 25.8 ± 0.5 N for 3 × 2 mm (Fig. 3d). The maximum bending angle also increases with the number of modules (Fig. 3e). Importantly, the peak bending force remains nearly constant across different module counts (~16–18 N, Fig. 3f). This reinforces the key advantage that the actuation force does not scale with the number of modules. As a result, we choose four modules with the 2 × 2 mm tendon for TeCoBot.

We next quantify the docking force, the maximum load before a connection fails, for different docking types (Fig. 3g–i). For claw-to-claw docking, we consider symmetric (i.e., 3-to-3) and asymmetric (i.e., 3-to-1) connections using latex and silicone sleeves of different lengths (Fig. 3g). Docking force is weakly affected by sleeve length but depends on material: latex achieves a higher mean peak force for symmetric docking (42.9 ± 4.5 N) than silicone (38.9 ± 2.6 N), whereas peak forces for asymmetric docking are comparable. For claw-to-body docking, we evaluate when a claw grasps a spacer disk (i.e., claw-plate) or a TPU tendon (i.e., claw-tensegrity). As the gripping force is proportional to the gripper motor current, the docking force increases with the current (max 1 A, Fig. 3h), reaching mean peak forces of 10.7/13.9 N (latex and silicone) on the plate and 29.8/16.0 N on the tensegrity. Finally, we evaluate the docking between a claw and a docking station, which replicates a claw finger without the sleeve. The docking force shows similar current-dependent trends (Fig. 3i), reaching 17.9 N (latex) and 19.7 N (silicone). Although silicone leads to stronger forces when docking to a station, we choose latex with medium sleeve length because it provides a much larger force for claw-tensegrity docking.

We develop a kinematic model based on constant curvatures[67] to predict TeCoBot's deformation under cable actuation, validated against experimentally reconstructed backbone shapes for four representative actuation patterns (Fig. 3j): tip position errors range from 1.9 to 6.4 mm (~0.8–3.0% of backbone length). Top-view visualizations (Fig. 3k) illustrate how individual cable pulls combine into a resultant vector that defines the bending-plane direction. Projecting the model-generated backbone onto the bending plane and overlaying it with two-dimensional images of the physical robot (Fig. 3l) reveals strong agreement between model arcs and experimental silhouettes. Detailed model formulation and validations are provided in Supplementary Material (Note 3, Fig. S7).

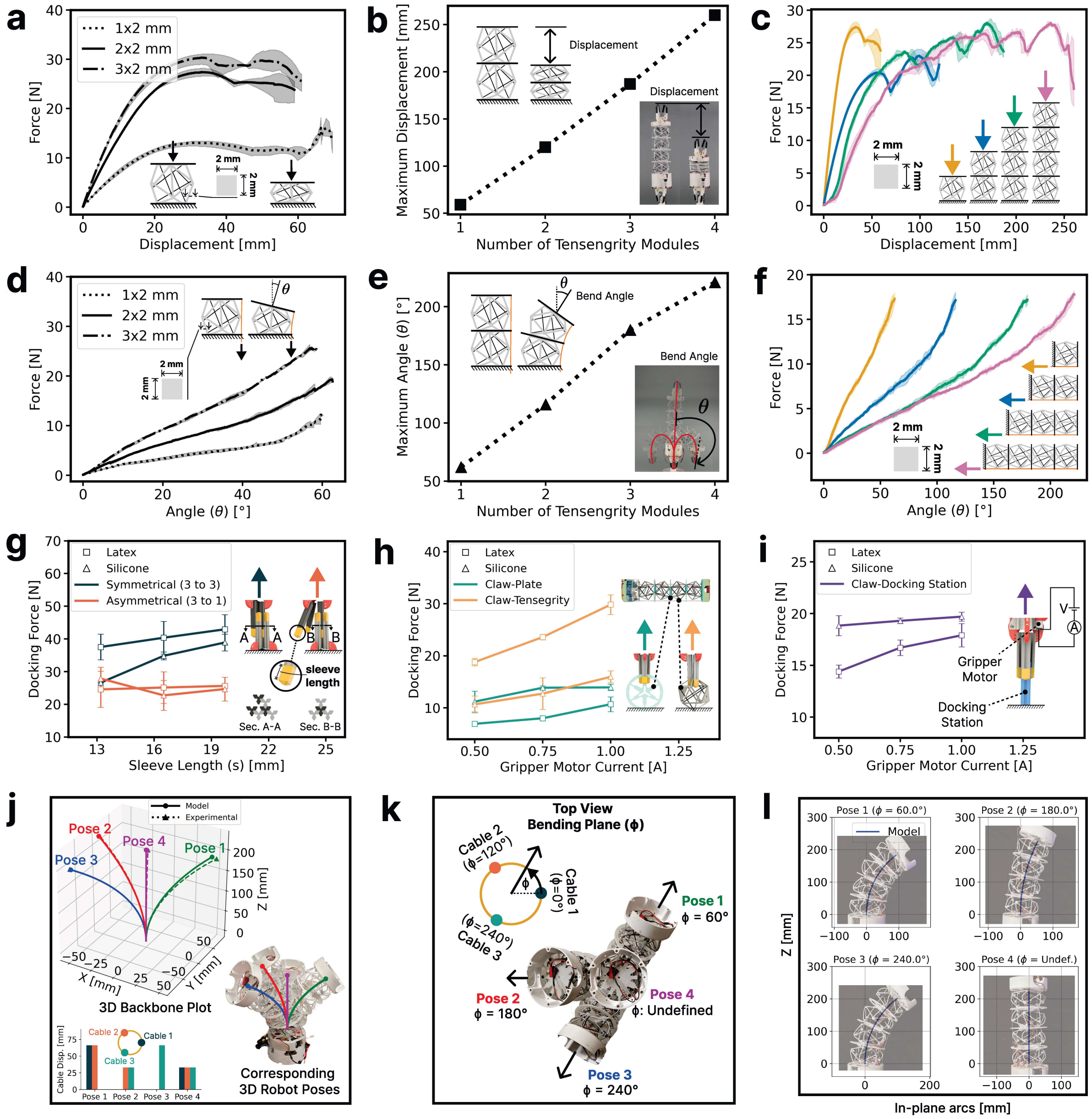

**Figure 3: Mechanical characterization and kinematic modeling of TeCoBot. (a)–(c), Axial collapsing characterization.** *(a)* Collapsing force–displacement curves for tendons with three cross-sectional sizes (1×2 mm, 2×2 mm, 3×2 mm), showing that thicker tendons require higher actuation force with diminishing increase beyond the intermediate size. *(b)* Maximum axial displacement increases linearly with the number of stacked tensegrity modules, while *(c)* the peak collapsing force remains approximately constant from one to four modules, demonstrating scalable deformation without increased actuation effort. **(d)–(f), Bending characterization.** *(d)* Bending force increases with tendon thickness for comparable bend angles. *(e)* Maximum achievable bending angle increases with module count. *(f)* Peak bending force is nearly independent of the number of modules, reinforcing that actuation force does not scale with the number of modules. **(g)–(i), Docking performance of claw-based end effectors.** *(g)* Claw-to-claw docking forces for symmetric and asymmetric engagements across sleeve materials and lengths, showing limited sensitivity to sleeve length but significant dependence on material choice. *(h)* Claw-to-body docking forces increase monotonically with motor current, with latex sleeves and tendon-based attachment yielding the highest forces. *(i)* Claw-to-docking-station interactions show similar trends, confirming robust attachment across

docking types and materials. **(j)–(l), Kinematic modeling and validation.** *(j)* Constant-curvature (CC) model predictions of backbone deformation closely match experimentally reconstructed shapes under four representative cable-actuation patterns. *(k)* Top-view comparison of model-predicted bending-plane directions with physical robot poses. *(l)* Projections of model backbones onto bending planes overlay closely with two-dimensional silhouettes of the robot. Quantitative validation results, including validations of the model parameters, are presented in Supplementary Materials. For panels (a), (c), (d), and (f)–(i), experiments were repeated three times. Solid lines or points represent the average across three trials, while shaded regions and error bars represent the measured range across the three trials. Panels (b) and (e) report the average values extracted from these repeated measurements.

## Manipulation and Locomotion with a Single TeCoBot

A single TeCoBot can grasp objects and locomote through a variety of modes by leveraging its compliant body and two claws to interact with objects and the environment. For manipulation (Supplementary Note 4, Movie S1), TeCoBot can axially extend/collapse and bend omnidirectionally to generate a three-dimensional workspace, as calculated from the model (Supplementary Fig. S8) and demonstrated experimentally (Supplementary Fig. S10a–c). These demonstrations include pick-and-place tasks, such as grasping a grape and placing it into a jar (Supplementary Fig. S10d–e). The claws are modular and can be swapped to match task requirements, expanding the range of objects and interactions it can handle.

A single TeCoBot can also locomote in diverse modes (Supplementary Note 5, Movie S2), including axial extension–compression, inchworm, crawling, rolling, turning, and pipe climbing. Timestamped image sequences illustrate each mode (Supplementary Fig. S11a-f), and an obstacle course (pipe, ramp, and table) further demonstrates that it can switch between modes to navigate complex environments (Supplementary Fig. S12). We also quantify TeCoBot's outdoor performance for two representative modes, axial extension–compression and rolling, across eight terrains (Supplementary Figs. S13–S15, Movie S3). Axial extension–compression is reliable on rigid ground (5 mm/s on smooth wood) but becomes ineffective on soft or loose terrains (e.g., grass, hay, gravel). In contrast, rolling remains effective on unstructured substrates such as rocks and hay and achieves higher speeds on most terrains (maximum 30 mm/s on smooth wood). We further quantify the cost of transport (CoT) on smooth wood, obtaining a CoT of 7.4 for rolling, substantially lower than most representative modular robots (Supplementary Fig. S17). We also show that the locomotion speed can be improved with better motors without changing the robot's morphology (Supplementary Fig. S19, Movie S3).

## Self-Reconfigurable Cooperative Manipulation with Multiple TeCoBots

Each with its own manipulation and locomotion capability, multiple TeCoBots can self-reconfigure into task-adaptive morphologies for cooperative manipulation (Supplementary Note 6, Movies S4 and S5). We construct a cuboidal frame with docking stations along its edges (Supplementary Fig. S20), allowing each robot to anchor to these docking stations in different orientations. With this setup, many configurations are possible; we focus on four representative three-robot morphologies (M1–M4 in Fig. 4a): three robots docked vertically on the frame (M1), two vertical and one horizontal on the ground (M2), one vertical with two horizontal (M3), and three connected in series (M4).

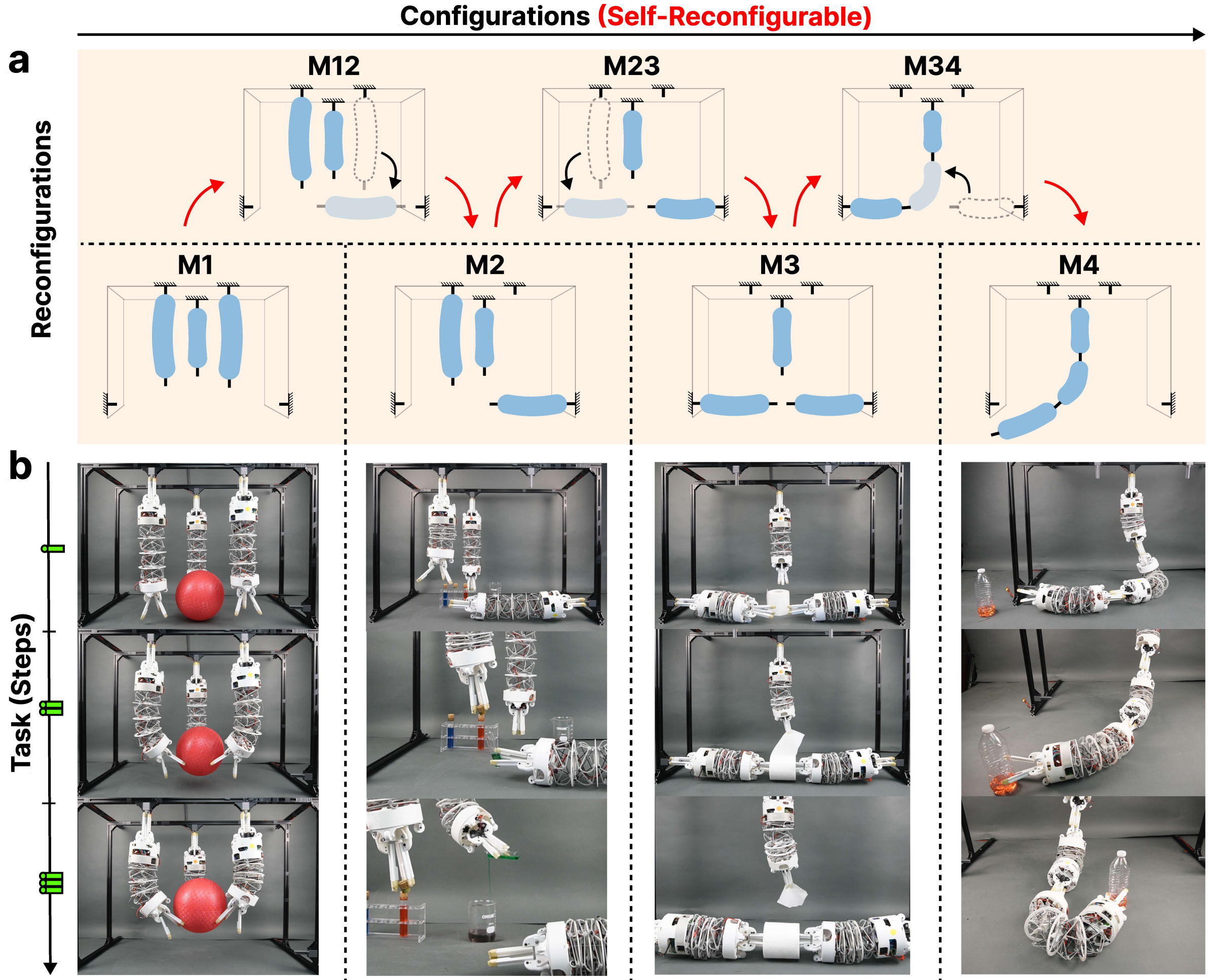

**Figure 4: Self-reconfigurable cooperative manipulation using multiple TeCoBots. (a)** Reconfiguration pathways between representative three-robot morphologies. TeCoBots can self-reconfigure between four different configurations (M1–M4) by sequential claw detachment, repositioning, and redocking. Transitions M12, M23, and M34 illustrate reconfiguration from vertical triple-robot docking (M1), to two-vertical-one-horizontal (M2), to one-vertical-two-horizontal (M3), and finally to a fully serial chain (M4). Dotted and solid bodies indicate previous and updated configurations, respectively, with arrows showing robot motion during the reconfiguration steps. **(b)** Cooperative manipulation demonstrations corresponding to the four configurations. In M1, three vertically docked robots bend inward to lift a large spherical object. In M2, the configuration supports multi-step manipulation tasks such as grasping a test tube, removing a cork, and pouring liquid into a container. In M3, two ground robots stabilize a tissue roll while the vertical robot tears off a piece. In M4, the serial morphology extends workspace reach, enabling the robots to retrieve and transport a bottle placed at a distance.

TeCoBots can self-reconfigure between these morphologies (Fig. 4a) via transitions M12, M23, and M34, which depict reconfiguration from M1 to M2, M2 to M3, and M3 to M4, respectively. For example, in M12, one robot undocks from the frame, lands on the ground, and docks horizontally, while the other two remain vertically docked. In M34, the robots can reconfigure into the serial arrangement by sequentially connecting to other robots and undocking from the frame. These transitions are also reversible. For instance, reconfiguring from M3 back to M2, along with all the reconfiguration processes, is demonstrated in Supplementary Fig. S21 and Movie S4.

Using these morphologies, we demonstrate cooperative manipulation of objects with different sizes and task requirements (Fig. 4b, Supplementary Movie S5). With M1, the three robots bend inward simultaneously to lift a large ball (200 mm diameter). With M2, the robots execute a multi-step task involving picking up a test tube, removing the cork, and pouring liquid into a beaker. With M3, two ground robots stabilize a tissue roll while the vertical robot unrolls and tears a section; this morphology also enables cooperative jar-opening (Supplementary Fig. S22c). With M4, the serial configuration extends the system's reach to access distant objects unreachable by a single robot. Additional cooperative manipulation behaviors, including those using other three-robot morphologies as well as two- and four-robot morphologies, are shown in Supplementary Fig. S22 and Movie S5.

## Self-Reconfigurable Cooperative Locomotion with Multiple TeCoBots

Multiple TeCoBots can also self-reconfigure into task- and environment-specific morphologies for cooperative locomotion (Supplementary Note 7, Movies S6 and S7). We highlight four representative three-robot configurations (L1–L4, Fig. 5) capturing distinct reconfiguration pathways, locomotion behaviors, and underlying mechanisms. Specifically, L1 forms a serial chain (---), L2 a loop with a trailing robot (O-), L3 an H-shaped morphology with two parallel robots connected by a central robot (H), and L4 a closed loop (O).

To reconfigure between morphologies, a TeCoBot can detach, move, and reattach to another TeCoBot's claw or body (Fig. 5a). For example, from L1 (---), one robot detaches while the other two form a loop, then the detached robot reconnects to the loop via claw-to-body connection to form L2 (O-). The reconfiguration pathways are reversible; rather than a direct transition from L3 (H) to L4 (O), the system can reconfigure through intermediate L2 (O-) and L1 (---) configurations before reconnecting to form the closed loop. These self-reconfigurations are demonstrated in Supplementary Fig. S23 and Movie S6.

These configurations enable cooperative locomotion in different environments (Fig. 5b and Supplementary Movie S7 for experimental demonstrations, Fig. 5c for the locomotion mechanisms). For L1 (---), the middle robot drives body extension and contraction while the front and rear robot claws alternately anchor by friction. For L2 (O-), the looped robots enclose a large object, while the trailing robot advances using an inchworm gait. For L3 (H), the system can traverse rectangular ducts, where the parallel robots alternately anchor to the duct, while the central robot advances through coordinated push–pull cycles. For L4 (O), all robots can roll in coordination to advance along large cylindrical objects.

Beyond three-robot configurations, we also demonstrate cooperative locomotion with two and four TeCoBots (Supplementary Fig. S24, Movie S7). Two robots in a linear '--' configuration can roll across gaps that are impassable for a single robot, while four robots in 'H-' and □ (square) morphologies enable lizard-like locomotion and collective forward translation as a unified body, respectively.

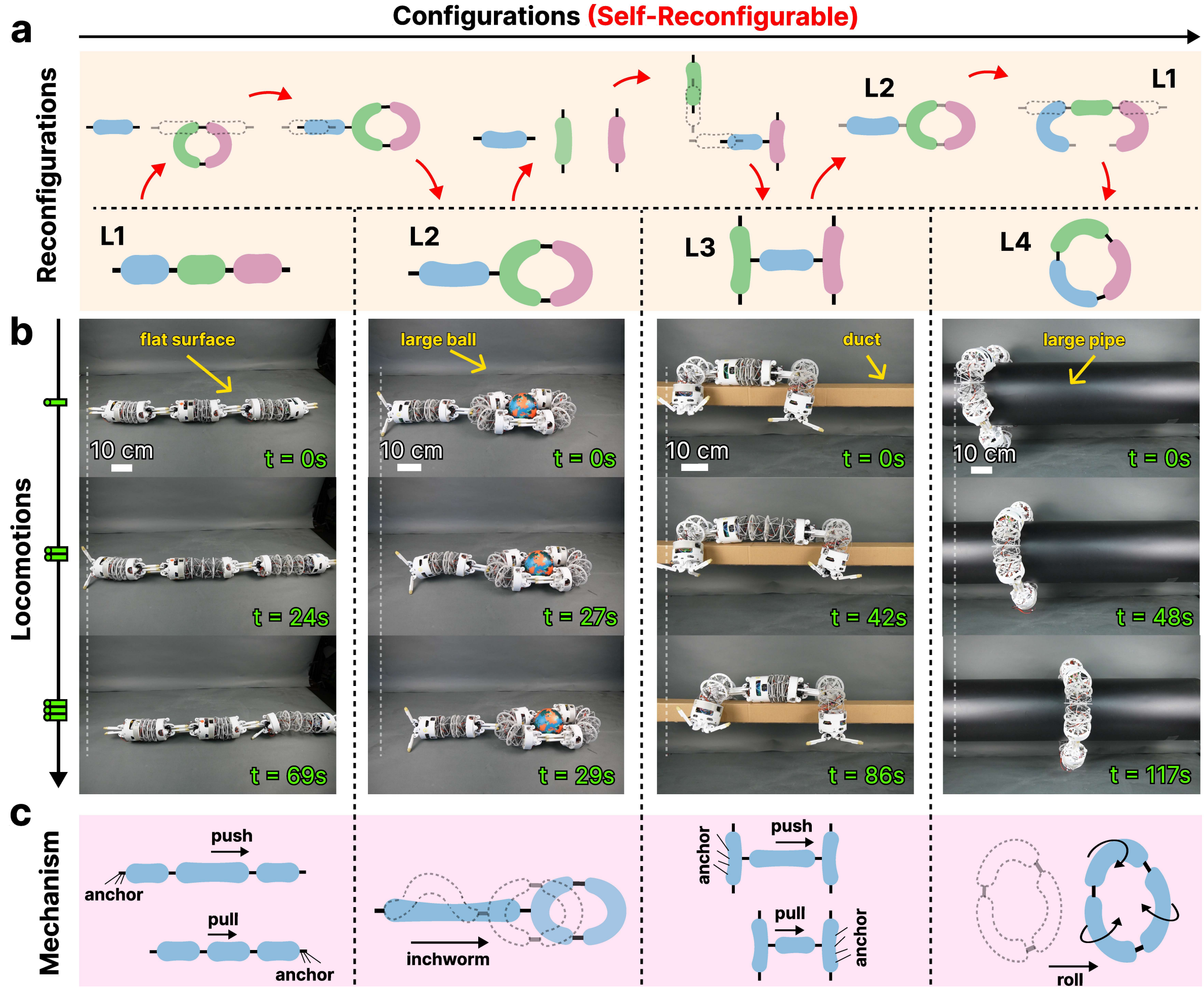

**Figure 5: Self-reconfigurable cooperative locomotion using multiple TeCoBots. (a)** Reconfiguration pathways among four representative locomotion morphologies (L1–L4), enabled by claw-to-claw and claw-to-body docking. Robots transition between linear (L1), loop-with-trailer (L2), H-shaped (L3), and closed-loop (L4) assemblies by sequential docking and undocking. **(b)** Locomotion demonstrations for each morphology: anchored extension–contraction on flat ground (L1), object-assisted inchworm motion while enclosing a large ball (L2), push–pull locomotion outside of a duct (L3), and rolling around a large cylindrical pipe (L4). **(c)** Schematic mechanisms underlying each locomotion mode, highlighting coordinated anchoring, pushing–pulling, inchworm, and loop-based rolling behaviors. In both (a) and (c), dotted robot bodies indicate the previous state, while solid bodies represent the subsequent state; distinct colors are used to differentiate the three robots during reconfiguration.

### Self-Reconfigurable Cooperative Loco-Manipulation

Building on the cooperative manipulation and locomotion capabilities, we demonstrate combined loco-manipulation in two real-world indoor scenarios, where three TeCoBots self-reconfigure to accomplish multi-phase objectives involving climbing, reaching, handover, and transport (Fig. 6, Supplementary Movies S8 and S9). Two additional outdoor demonstrations are provided in Supplementary Note 9, Figs. S25-S26, Movie S10.

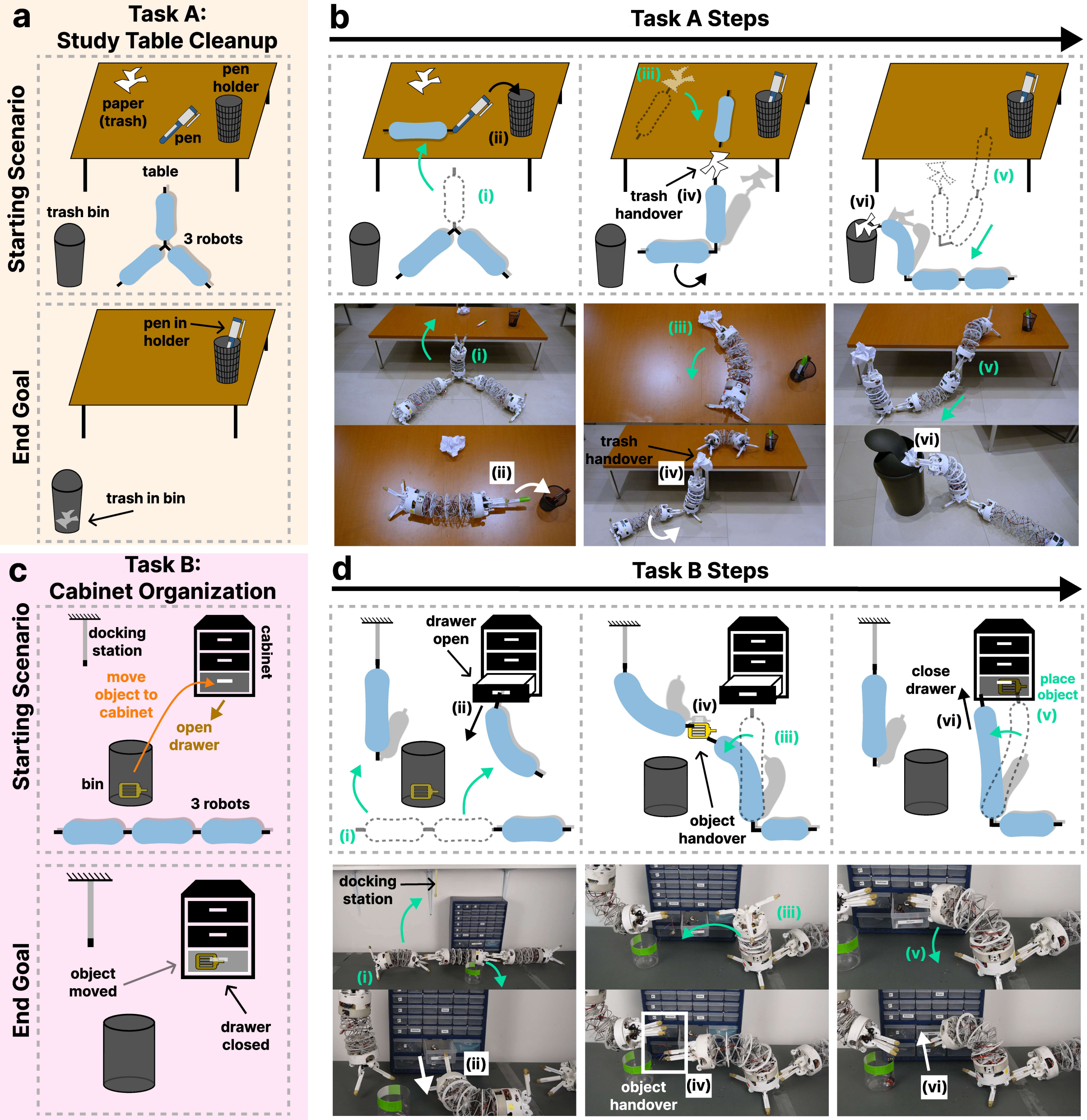

**Figure 6: Self-reconfigurable cooperative loco-manipulation with multiple TeCoBots.** For each task, the leftmost column illustrates the starting scenario and end goal to convey the task objective. The top rows show schematic task progression, while the bottom rows present the corresponding physical robot executions. Each task is illustrated using three columns, with two steps per column (six steps total, labeled i–vi). In the schematics, dotted robot bodies denote the previous state, and solid bodies denote the subsequent state. Color is used exclusively to distinguish step order and motion direction (turquoise for the first step, black/white for the second). Shadows are applied only to robot bodies to indicate three-dimensional posture, such as standing or hanging vertically and lying horizontally. **(a–b) Task A: Study table cleanup.** Three TeCoBots self-reconfigure through a sequence of standing, climbing, handover, and serial-chain formation to return a pen to its holder and dispose of scrap paper. Schematics (top) show key reconfiguration and manipulation steps (i–vi), and photographs (bottom) show corresponding physical executions, including climbing onto the table, grasping and transferring the paper, forming a three-robot chain, and depositing the trash into the bin. **(c–d) Task B: Cabinet organization.** The robots autonomously transition between vertical docking, serial connection, and cooperative handover to retrieve an object from a bin and place it into a drawer. Schsematics (top) illustrate the stepwise sequence—opening the drawer, object retrieval, vertical-to-ground handover, and drawer closing—while photographs (bottom) show the physical execution of each phase.

### Task A: Study Table Cleanup

In Task A, the robots need to return a misplaced pen to its holder on a table (height: 40 cm) and dispose of scrap paper into a trash bin (Fig. 6a). Starting from the floor, the task requires the robots to climb onto the table, fetch the paper, and descend to discard it.

The robots follow several major steps to accomplish the task (Fig. 6b). Starting from a 'Y' configuration on the floor, the tail robot reorients vertically while the two base robots push it onto the table (step i). After that, it moves to place the pen into the holder and grasps the scrap paper (steps ii-iii). The two ground robots then form a serial chain, and one stands up to receive the paper via cooperative handover (step iv). The other ground robot then rotates and aligns to dock with the robot on the table, forming a three-robot chain spanning the table and floor (step v). This connected assembly advances toward the trash bin, where the robot carrying the paper reorients vertically to dispose of it (step vi). The task demonstrates coordinated climbing, manipulation, handover, and reconfiguration (Supplementary Movie S8).

### Task B: Cabinet organization

Task B involves organizing components between a supply bin and a cabinet drawer (Fig. 6c), aided by a docking station on a wall-mounted shelf adjacent to the compartment box. The robots conduct the following steps to accomplish the task (Fig. 6d). Starting with a serial (---) configuration on the ground, one robot stands up and docks to the station while another opens the drawer (steps i–ii). The docked robot then retrieves a component (motor) from the bin. Meanwhile, the two undocked robots form a series chain (--), and one reorients vertically to receive the motor through cooperative handover (steps iii–iv); then it places the motor into the drawer and closes it by pushing, while remaining connected to and supported by the base robot (steps v–vi). Additional retrieval and placement steps after reaching the goal are omitted for clarity, and the complete sequence is shown in Supplementary Movie S9.

## DISCUSSION

This work presents a design framework that combines local body compliance with global, multi-robot reconfiguration to enable rich task-adaptive morphologies for cooperative behaviors. We instantiate this framework using TeCoBot, a tensegrity-based continuum robot with dual claws. Each TeCoBot can bend omnidirectionally and collapse/extend axially for locomotion and grasping. It can also connect with other units to form different morphologies for complex cooperative behaviors, such as coordinated manipulation, lifting, object transport, climbing, and adaptive locomotion. These behaviors are demonstrated in real-world scenarios, including workspace cleanup, cabinet organization, and outdoor tasks, highlighting its adaptability across complex loco-manipulation tasks.

These two properties (local body compliance and global multi-robot reconfiguration) are typically achieved only separately: rigid MRRs achieve global reconfiguration but lack local compliance, while continuum robots provide compliance but generally lack multi-robot reconfiguration. TeCoBot exhibits both properties: its tensegrity-based body provides compliance, while its dual claws enable global reconfiguration. Compliance, in turn, enables embodied intelligence, offloading part of the sensing and control burden onto the body itself. TeCoBot's cable-strut tensegrity backbone passively conforms around objects, tolerates docking and contact misalignment, and distributes forces across uneven terrain. Rigid MRRs lack this embodied intelligence, since they must actively sense and act to achieve a similar level of interaction. Global reconfiguration, meanwhile, comes from claws at both ends, which enable mechanically stable claw-to-claw and claw-to-body docking to form chains, loops, and branched assemblies. Since the connected modules remain deformable, these assemblies can also bend and collapse/extend, enabling rolling, inching, turning, and pipe climbing. As the number of TeCoBots increases, the assembly space expands not only through additional docking combinations but also through the continued deformation of each connected module (Supplementary Note 10, Fig. S27), enabling the same robots to function as manipulators, mobile platforms, or structural supports without hardware modification. It is the combination of both properties—compliance from tensegrity, reconfiguration from modular docking—that enables TeCoBot's task-adaptive capabilities.

While some locomotion modes are automated (e.g., rolling, crawling), the self-reconfigurations are teleoperated by human operators, allowing us to focus on the system's cooperative behavioral capabilities. Nevertheless, docking-based self-reconfigurations can be automated, as demonstrated through both open-loop autonomous docking with pre-scripted motions and closed-loop autonomous docking using onboard vision and fiducial-marker feedback (Supplementary Note 11, Movie S11). These results show that TeCoBot can autonomously approach, align with, and dock to another robot using the existing platform with minimal hardware modification. Based on these results, a promising direction for future work would be developing autonomous decision-making frameworks capable of selecting optimal morphologies based on task requirements, environmental context, and system constraints[68]. These capabilities open pathways to wide applications ranging from inspection and transport in logistics and manufacturing to adaptive navigation in search-and-rescue or space exploration, where robots must operate in uncertain and highly constrained environments.

## METHODS

### Robot Design and Fabrication

Each TeCoBot consists of vertically stacked tensegrity modules that form a compliant continuum structure. A single module is based on an icosahedral tensegrity geometry, composed of six rigid carbon fiber rods (58 mm in length, 2 mm in diameter) and a continuous, planar tendon network made from thermoplastic polyurethane (TPU) (Fig. 2b). The tendon structure is fabricated using fused deposition modeling (FDM) 3D printing and assembled into the 3D tensegrity form by inserting rods into pre-defined anchor points on the tendon network.

The base, top module, and spacer disks are 3D-printed with polylactic acid (PLA) material. The parts are fastened with M3 screws (3 mm diameter) and threaded heat-set inserts. Three spacer disks (110 mm diameter, 2 mm thickness) are placed between adjacent modules to guide internal actuation cables and maintain structural alignment. Each robot includes three actuated cables routed through the body and controlled by encoded DC motors (Pololu 380:1 Micro Metal Gearmotor) mounted at the base. Motors are arranged 120° apart and equipped with encoders for closed-loop bidirectional control of cable spooling. The cable spool and motor cover are 3D printed with polycarbonate to provide higher strength and durability. The actuation cable is a 40 lb. braided fishing line, which is anchored at the top module secured with screws.

### Control and Electronics

Each TeCoBot is wirelessly controlled via Wi-Fi using ESP32 microcontrollers. A central ESP32-S3 transmitter connected to a PC sends commands to an onboard ESP32-S3 receiver on each robot (Fig. 2e). The base module of each robot houses the ESP32-S3 board and three motor drivers (Dual TB6612FNG, SparkFun). The system is powered by two Lithium-Polymer batteries: one at the base module (2 cells, 300 mAh) for the controller and another at the top module (3 cells, 450 mAh) for the motors. This setup allows both synchronized and untethered control of multiple connected robots.

**Claw and Roller Wheel Design**

Two interchangeable claw designs are used depending on the task. The first uses 3D-printed (PLA) fingers fitted with a rubber tubular sleeve at its end (Supplementary Fig. S2g-i) to enhance friction for locomotion and multi-point docking. The second design integrates a soft 3D-printed TPU pad on a PLA finger base, optimized for manipulating soft or delicate objects with limited connection capability. Each claw is actuated in one of two ways: using either a smart servo (Dynamixel XL330-M288-T, Robotis) or a geared DC motor (380:1 Micro Metal Gearmotor, Pololu), driving a pinion gear that ultimately drives a worm gear (Fig. 2d) which self-locks to hold the grip without power. The hex and the gears are all 3D-printed with PETG (Polyethylene Terephthalate Glycol) material. In addition to the grippers, 3D-printed roller wheels with PLA materials are mounted in between each finger of the claws.

**Characterization Experiments**

We evaluate the mechanical response of the tensegrity modules for both collapsing and bending deformations (Supplementary Fig. S1a-b). For collapse tests, a cable is mounted at the midpoint of the tensegrity module stack, and a force gauge (M3-20, Mark-10, 100 N range) applies a vertical pulling force through a custom test rig (Fig. S1a). The force and displacement are recorded simultaneously during actuation. Bending tests are conducted with the same test rig, with the actuation cable attaching to the edge of the top tensegrity module while the base remains fixed. We measure the angle using a digital protractor (XRCLIF Magnetic Angle Gauge, 4 × 90° range). Modules with tendon sizes 1×2 mm, 2×2 mm, and 3×2 mm are tested (Supplementary Fig. S1c). Supplementary Note 2 details the experimental procedures for characterization.

For docking characterization, we evaluate the maximum holding force generated by the claw-based end effectors under different docking configurations and materials (Supplementary Fig. S2b-f). In these experiments, claws equipped with a smart servo motor (Dynamixel XL330-M288-T, Robotis) are exclusively used (Supplementary Fig. S2a) for controlling the gripping force. Three docking types are tested: claw-to-claw, i.e., symmetrical and asymmetrical docking (Supplementary Fig. S2b-c), claw-to-body, i.e., claw-plate and claw-tensegrity docking (Supplementary Fig. S2d-e), and claw to a docking station (Supplementary Fig. S2f). For all tests, elastomer sleeves made of latex and silicone are mounted on the claw fingers. Docking force is measured using the same force gauge while holding the motor current steady at a predetermined value until detachment, which ensures similar levels of gripping force across trials.

For the kinematic model validation, we construct physical robot shapes using images from experiments and compare against model-predicted backbones computed from cable displacements. We validate the model systematically by measuring the experimental backbone effective length, bending angle, and bending-plane direction under different cable actuation patterns (Supplementary Fig. S7a-c). Two inertial measurement units (BNO085, Adafruit) are mounted on the base and top plates of the robot to estimate bending angle and bending plane (Supplementary Fig. S5). A time-of-flight sensor (VL6180X, Adafruit) is attached to a steel wire guided through the middle of the robot to measure the length change and effective length of the robot backbone (Supplementary Fig. S6). Supplementary Note 3 details the model formulation and systematic validation along with model workspace computation and inverse kinematics (Supplementary Figs. S7-S9).

### Locomotion and Manipulation Experiments

We build a frame with aluminum extrusions (20 mm × 20 mm) for manipulation tasks (Supplementary Fig. S20). We then equip the frame with customized docking stations (3D-printed with PLA) placed along its edges. These docking stations enable TeCoBots to anchor in various configurations and perform cooperative manipulation tasks (Fig. 4 and Supplementary Movie S5). Indoor locomotion experiments are performed on a flat table surface covered with a sheet of background paper for visual contrast (Fig. 5 and Supplementary Movie S6). No external fixtures are used for locomotion. A series of pre-defined motor commands are used to generate certain locomotion modes (Supplementary Fig. S16). We teleoperate all the TeCoBots wirelessly for all demonstrations except scripted locomotion and autonomous docking. All details for manipulation and locomotion experiments are provided in Supplementary Notes 4, 5, 6, 7, and 8. Additional loco-manipulation experiments in outdoor environments are detailed in Supplementary Note 9.

### Cost of Transport

We quantify the cost of transport (CoT) for axial extension–compression and rolling locomotion on smooth wood, powering the robot with an external DC power supply and measuring the average electrical power P over one locomotion cycle. Following the standard definition, CoT = P/(mgv), where m is the robot mass, g is gravitational acceleration, and v is the measured average forward speed. Because smooth wood minimizes slip and terrain-induced losses, the resulting CoT values represent a lower-bound estimate; full measurement details are provided in Supplementary Note 5.3.

### Locomotion Speed Improvement Analysis

The TeCoBot prototype is designed to validate compliant modular reconfiguration rather than to optimize locomotion speed, and its axial extension–compression cycle is dominated by slow claw actuation (~7–10 s per open/close). We can increase the locomotion speed from ~5 mm/s to ~12 mm/s (a 2.4× improvement) on smooth wood by replacing the claw's 380:1 Pololu gearmotor with a faster 30:1 motor of the same locking-worm-gear design without changing the robot's morphology or gait. We can potentially further increase the overall speed by ~5.5× by replacing the current body motor with one (e.g., Dynamixel XL330-M077-T, Robotis) having a greater no-load speed. Crawling-with-claws locomotion would also benefit from the faster claw motors, increasing the estimated speed from ~2.25 mm/s to ~6.1 mm/s (~2.7× improvement). Rolling speed can also be improved using closed-loop IMU feedback which can reduce conservative open-loop timing. Full derivations, motor-command timing, and projected speeds are provided in Supplementary Note 5.4 (Figs. S18–S19, Movie S3).

### Autonomous Docking

Beyond teleoperation, we test open-loop and closed-loop autonomous docking. In open-loop docking, a mobile TeCoBot executed a pre-scripted sequence of rolling, alignment, and claw-actuation commands to complete a 3-to-3 claw-to-claw docking with a stationary robot without visual feedback. For closed-loop docking, we added an autonomy module—a Raspberry Pi 5, a Pi camera, an ESP32 transmitter, and a Dynamixel smart-servo front claw—inside the mobile robot's top module, without modifying the existing motor-control architecture. The camera detects two ArUco markers on the stationary robot: a large marker for coarse, long-range alignment and a small marker for close-range docking assessment. The controller computes the image-space marker error and applies threshold-based lateral corrections, alternating axial extension-compression motion with re-alignment checks (a sense–correct–move cycle) until a marker-size and alignment criterion indicates the robot is docking-ready, at which point the front claw closes using a current-aware stopping routine. Full hardware details and the control-flow algorithm are provided in Supplementary Note 11 (Figs. S28–S29, Movie S11).

### Software

Force–displacement data were recorded using a Mark-10 force gauge with MESUR Lite 2.0.1. Robot firmware was developed in Arduino IDE 2.2.1 using the ESP32 Arduino core 2.0.16 (Espressif Systems), with the Adafruit NeoPixel 1.12.2 and ESP32Encoder 0.11.6 libraries. DYNAMIXEL servos were commanded using the DYNAMIXEL SDK (dynamixel-sdk 3.7.31). For the autonomous docking experiments, video was captured on a Raspberry Pi running Raspberry Pi OS (Debian 13) using rpicam-vid (rpicam-apps 1.12.0, libcamera 0.7.1) and processed in real time using OpenCV 4.13.0. Measurements were recorded in Microsoft Excel Version 2608 (Microsoft 365). Data analysis and figure generation were performed in Python 3.13.5 using NumPy 2.1.3, pandas 2.2.3, Matplotlib 3.10.0 and scikit-learn 1.6.1.

## Data Availability

Source data for Figures 1 and 3 are provided with this paper. The source data underlying the figures are available on figshare at https://doi.org/10.6084/m9.figshare.33578671[69]. File-by-file descriptions, including column definitions and units, are provided in the README accompanying the deposit.

## Code Availability

The custom code developed for this study, including firmware for the ESP32 transmitter and receiver units, the constant-curvature validation data collection scripts, and the autonomous visual docking implementation, is publicly available at https://github.com/mahmud-hasans/TeCoBot under an MIT license. The specific version used to generate the results reported in this paper is archived on figshare at https://doi.org/10.6084/m9.figshare.33505864[70].

## Acknowledgements

This work is partially supported by the National Science Foundation (NSF) under grant CNS – 2334883, and the Advanced Industries Proof of Concept Grant from the Colorado Office of Economic Development and International Trade (OEDIT).

## Supplementary Materials

Supplementary Notes 1 to 11
Supplementary Figures S1 to S29
Supplementary Tables S1 to S3
Supplementary Movies S1 to S11